\documentclass[10pt,twocolumn,letterpaper]{article}

\usepackage{cvpr}              

\usepackage{graphicx}
\usepackage{amsmath}
\usepackage{amssymb}
\usepackage{booktabs}
\usepackage{xcolor,colortbl}
\usepackage[title]{appendix}
\usepackage{adjustbox}

\usepackage{subfiles}
\usepackage{caption}
\DeclareCaptionLabelFormat{AppendixTables}{A.#2}

\usepackage[pagebackref,breaklinks,colorlinks]{hyperref}

\usepackage[capitalize]{cleveref}
\crefname{section}{Sec.}{Secs.}
\Crefname{section}{Section}{Sections}
\Crefname{table}{Table}{Tables}
\crefname{table}{Tab.}{Tabs.}

\def\cvprPaperID{1670} 
\def\confName{CVPR}
\def\confYear{2023}

\DeclareCaptionLabelFormat{AppendixTables}{A.#2}

\begin{document}

\title{ABLE-NeRF: Attention-Based Rendering with Learnable Embeddings for Neural Radiance Field}
\author{Zhe Jun Tang\textsuperscript{1} \qquad
Tat-Jen Cham\textsuperscript{2} \qquad
Haiyu Zhao\textsuperscript{3}
\and
\textsuperscript{1}{S-Lab, Nanyang Technological University} \qquad
\textsuperscript{2}{Nanyang Technological University} \\
\textsuperscript{3}{SenseTime Research} 
}

\maketitle

\begin{abstract}

    Neural Radiance Field (NeRF) is a popular method in representing 3D scenes by optimising a continuous volumetric scene function. Its large success which lies in applying volumetric rendering (VR) is also its Achilles' heel in producing view-dependent effects. As a consequence, glossy and transparent surfaces often appear murky. A remedy to reduce these artefacts is to constrain this VR equation by excluding volumes with back-facing normal. While this approach has some success in rendering glossy surfaces, translucent objects are still poorly represented. In this paper, we present an alternative to the physics-based VR approach by introducing a self-attention-based framework on volumes along a ray. In addition, inspired by modern game engines which utilise Light Probes to store local lighting passing through the scene, we incorporate Learnable Embeddings to capture view dependent effects within the scene. Our method, which we call ABLE-NeRF, significantly reduces `blurry' glossy surfaces in rendering and produces realistic translucent surfaces which lack in prior art. In the Blender dataset, ABLE-NeRF achieves SOTA results and surpasses Ref-NeRF in all 3 image quality metrics PSNR, SSIM, LPIPS.
    
\end{abstract}

\section{Introduction}
\label{sec:intro}

Neural Radiance Field (NeRF) has become the de facto method for 3D scene representation. By representing the scene as a continuous function, NeRF is able to generate photo-realistic novel view images by marching camera rays through the scene. NeRF first samples a set of 3D points along a camera ray and outputs its outgoing radiance. The final pixel colour of a camera ray is then computed using volumetric rendering (VR) which colours are alpha-composited. This simple approach allows NeRF to generate impressive photo-realistic novel views of a complex 3D scene. However, NeRF is unable to produce accurate colours of objects with view-dependent effects. Colours of translucent objects often appear murky and glossy objects have blurry specular highlights. Our work aims to reduce these artefacts.

\begin{figure}[t]
\centering
\includegraphics[width=1.0\linewidth]{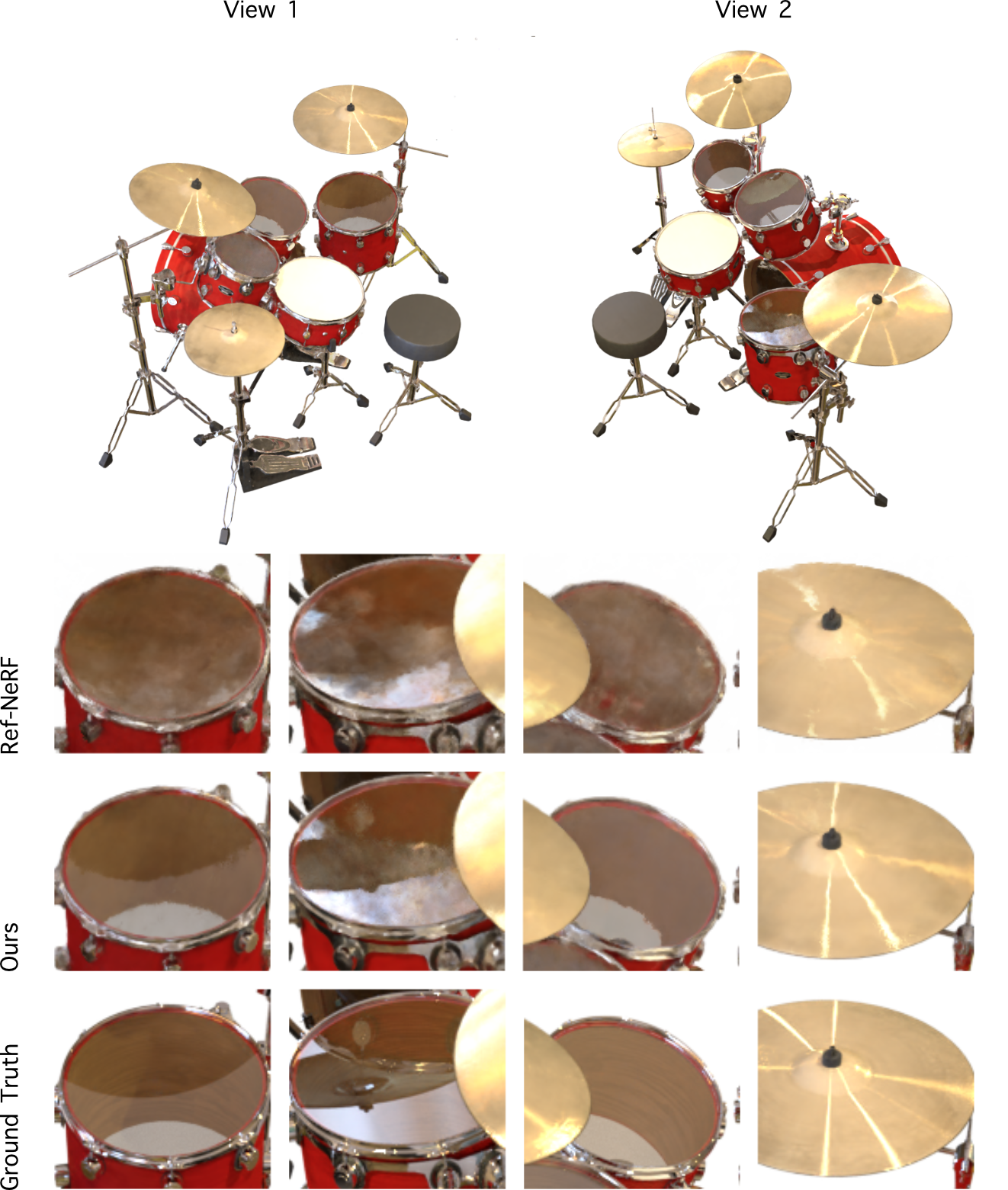}
\caption{We illustrate two views of the Blender 'Drums' Scene. The surface of the drums exhibit either a translucent surface or a reflective surface at different angles. As shown, Ref-NeRF model has severe difficulties interpolating between the translucent and reflective surfaces as the viewing angle changes. Our method demonstrates its superiority over NeRF rendering models by producing such accurate view-dependent effects. In addition, the specularity of the cymbals are rendered much closer to ground truth compared to Ref-NeRF.}
\label{fig:drums-1}
\end{figure}

The exhibited artefacts of the NeRF rendering model is largely due to the inherent usage of VR as features are accumulated in the colour space. Variants of NeRF attempt to tackle this defect by altering the basis of this VR equation. For instance, Ref-NeRF first predicts the normal vector of each point on the ray. If a point has a predicted normal facing backwards from the camera, its colour is excluded from computation via regularisation. However, prediction of normals in an object's interior is ill-posed since these points are not on actual surfaces. As a consequence, Ref-NeRF achieves some success over the baseline NeRF model, albeit imperfectly. 

When rendering translucent objects with \emph{additional specular effects}, NeRF and its variants suffer from the same deficiency. This is due to the computation of $\sigma$ which is analogous to the `opacity' attribute of a point used in the VR equation. It is also related to the point's transmissivity and its contribution of radiance of to its ray. As per the Fresnel effect \cite{born2013principles}, this property should depend on viewing angles. Similarly, \cite{alphasphere} describes a notion of \emph{`alphasphere'}, which describes an opacity hull of a point that stores an opacity value viewed at direction $\omega$. Most NeRF methods disregard the viewing angle in computing $\sigma$. In fig.~\ref{fig:drums-1}, the surface of the uttermost right drum in the Blender scene exhibits changing reflective and translucent properties at different viewing angles. Ref-Nerf and other variants, by discounting the dependency of $\sigma$ on viewing angle, may not render accurate colours of such objects.

Additionally, learning to model opacity and colour separately may be inadequate in predicting the ray's colour. Accumulating high-frequency features directly in the colour space causes the model to be sensitive to both opacity and sampling intervals of points along the ray. Therefore we rework how volumetric rendering can be applied to view synthesis. Inspecting the VR equation reveals that this methodology is similar to a self-attention mechanism; a point's contribution to its ray colour is dependent on points lying in-front of it. By this principle we designed ABLE-NeRF as an attention-based framework. To mimic the VR equation, mask attention is applied to points, preventing them from attending to others behind it.

The second stage of  ABLE-NeRF takes inspiration from modern game engines in relighting objects by invoking a form of memorisation framework called `baking'. In practice, traditional computer graphics rendering methods would capture indirect lighting by applying Monte Carlo path tracing to cache irradiance and then apply interpolation during run-time. Similarly, game engines would use lightmaps to cache global illumination for lower computational costs. For relighting dynamic objects, localised light probes are embedded in the scene to capture light passing through free space. At run-time, moving objects query from these light probes for accurate relighting. The commonality between all these approaches is the process of `memorising' lighting information and interpolating them during run time for accurate relighting. As such, we take inspiration from these methods by creating a memorisation network for view synthesis. Given a static scene, we incorporate Learnable Embeddings (LE), which are learnable memory tokens, to store scene information in latent space during training. Specifically, the LE attends to points sampled during ray casting via cross-attention to memorise scene information. To render accurate view dependent effects a directional view token, comprising of camera pose, would decode from these embeddings.

ABLE-NeRF provides high quality rendering on novel view synthesis tasks. The memorisation network achieves significant improvements in producing precise specular effects over Ref-NeRF. Moreover, by reworking volumetric rendering as an attention framework, ABLE-NeRF renders much more accurate colours of translucent objects than prior art. On the blender dataset, ABLE-NeRF excels both quantitatively and qualitatively relative to Ref-NeRF.

In summary, our technical contributions are:

(1) An approach demonstrating the capability and superiority of transformers modelling a physics based volumetric rendering approach.


(2) A memorisation based framework with Learnable Embeddings (LE) to capture and render detailed view-dependent effects with a cross-attention network.


\section{Related Work}
\label{sec:Related}

We first review techniques from computer graphics for capturing indirect lighting effects and global illumination. Following, we discuss how NeRF and its other variants render photo-realistic images of a 3D scene from an unseen viewing angle.

\textbf{Indirect Illumination in Rendering.}
Rendering with indirect illumination is a widely studied topic. Pioneering works using path tracing \cite{kajiya1986rendering} or light tracing \cite{dutre1995importance} cast rays from a camera until they hit a point and traces random rays at the visible surface to light sources. However, these methods requires heavy computation as sampling multiple rays is a costly operation. Instead, irradiance caching \cite{krivanek2009practical} is applied to sparsely samples rays and its indirect illumination is stored to speed up this process. An object's illumination will then be interpolated at its nearby cached values. Other methods involving a pre-computation based method like radiance transfer and lightmaps \cite{abrash2000quake}, first calculate the surface brightness and store it in texture maps for real time performance. Unlike lightmaps storing surface lighting information, light probes \cite{lightprobes} bake lighting information passing through the scene. During run time, dynamic objects would query from the nearest light probes for indirect lighting information. The use of probes can be similarly be extended to reflections. In game engines, reflection probes \cite{reflectionprobes} are made to capture images as cubemaps within the scene. These cubemaps are then utilised by objects with reflective materials to produce convincing reflections of the environment. 

The impetus to incorporate Learnable Embeddings in our work takes inspiration from how these light or reflection probes function. Yet, our work differs from the traditional graphics pipeline in the type of information being captured. Unlike probes in game engines, these embeddings do not exist as physical entities in the 3D scene geometry. Instead, Learnable Embeddings operate within the latent space as learnable latent vectors. In this manner, the LE capture latent information of a given scene. Thus, it is crucial to optimise these LE via training. Similar to relighting dynamic objects by interpolating nearby light probes or reflection probes, new viewing angles would query from these LE to achieve accurate view dependent effects.

\textbf{3D Scene Representation for View Synthesis}
Numerous methods have been proposed for generating new images of a scene using only a few captured images. Light field rendering methods \cite{lightfieldrendering,lumigraph} characterise the unobstructed flow of light passing through the scene as 4D function and slice this slab differently to generate new views. While these methods require a large number of light field samples to interpolate new views, recent deep learning-based methods \cite{LFN} only require sparse inputs. Separately, image based rendering methods \cite{ibrnet,lightfieldneural,deepblending,unstructuredlumi,geoblend} balance a set of weights heuristically or learned to blend nearby input images creating novel views. Scene representation methods also extend to volumetric methods by colouring voxel grids \cite{voxelcoloring} or reconstructing plenoxels\cite{plenoxels}. Methods involving neural networks are also capable of learning volumetric scene representation through gradient-based methods \cite{deepview, deepvoxels, mpi, voxoctrees}

The shift towards coordinate-based methods has shown a quantum leap in concise 3D scene representation. With a few layers of MLP, NeRF \cite{nerf} can map a continuous input of 3D coordinates to the scene geometry and appearance. NeRF can also be extended to dynamic scenes, avatar animations, and even scene editing. These algorithms, which model appearance, typically decompose scene materials into its BRDF properties \cite{zhang2021nerfactor}. As a result, they require strong physics based assumptions such as known lighting conditions or single type materials. On the contrary, Ref-NeRF \cite{refnerf} does not assume these precise physical meanings. This enables Ref-NeRF to avoid relying on such assumptions. Our work follows this school of thought. We do not assume a physics based learning approach as we replace volumetric rendering by an end to end deep learning methodology.

\textbf{Transformers for View Synthesis}
The use of transformers for view synthesis have gained popularity lately. IBR-Net \cite{ibrnet} applies a ray transformer to generate $\sigma$ values before using the VR equation to accumulate colours. In \cite{lightfieldneural}, the authors apply a two-stage transformer-based model to aggregate features along epipolar lines in each reference views and then combine these reference views with a view transformer. SRT \cite{SRT} extracts features from training images with a CNN and then apply transformers to aggregate features before using a target ray to query for a pixel colour. NeRF-in-detail \cite{nerfindetail} also uses a transformer to propose new sample points along a ray and then apply NeRF to generate a ray colour. Unlike ABLE-NeRF, none of these methods apply transformers to model a physics based volumetric rendering approach.

\subsection{Neural Radiance Field Overview}
NeRF represents a 3D scene as a continuous volumetric scene function. It traces a pixel ray $\mathbf{r}(t) = \mathbf{o} + t\mathbf{d}$, into a scene where $\mathbf{o}$ and $\mathbf{d}$ represent the camera origin and pose. After sampling for 3D points along the ray, NeRF predict point's opacity using spatial MLPs. Following which, a directional MLP determines the colour of the point. Finally, to compute the colour of a ray, alpha composition with numerical quadrature is applied to these points based on (\ref{eqn:color}). 
\begin{equation} \label{eqn:color} 
\mathbf{\hat{C}}\left(\mathbf{r}\right) = \sum_{i=1}^{N}T_i\left(1-\exp{\left(-\sigma_i\delta_i\right)}\right)\mathbf{c}_i
\end{equation} 
where 
\begin{equation} \label{eqn:weight} 
T_i=\exp{\left(-\sum_{j=1}^{i-1}\sigma_j\delta_j\right)}
\end{equation} 

NeRF maintains two separate sets of model parameters for the coarse and fine network. The network is optimised with a total squared error between the predicted pixel colour and the true pixel colours of both the coarse and fine network.

\begin{equation} \label{eqn:loss} 
\mathcal{L} = \sum_{\mathbf{r}\in\mathcal{R}} \left[ \left\| \hat{C}_c \left( \mathbf{r} \right)-C_c \left( \mathbf{r} \right) \right\|^2_2 +  \left\| \hat{C}_f \left( \mathbf{r} \right)-C_f \left( \mathbf{r}\right) \right\|^2_2 \right]
\end{equation} 

In practise, only the output of the fine network is used to render the final image.

\section{Method}

\begin{figure*}[t]
\centering
\includegraphics[width=1.0\linewidth]{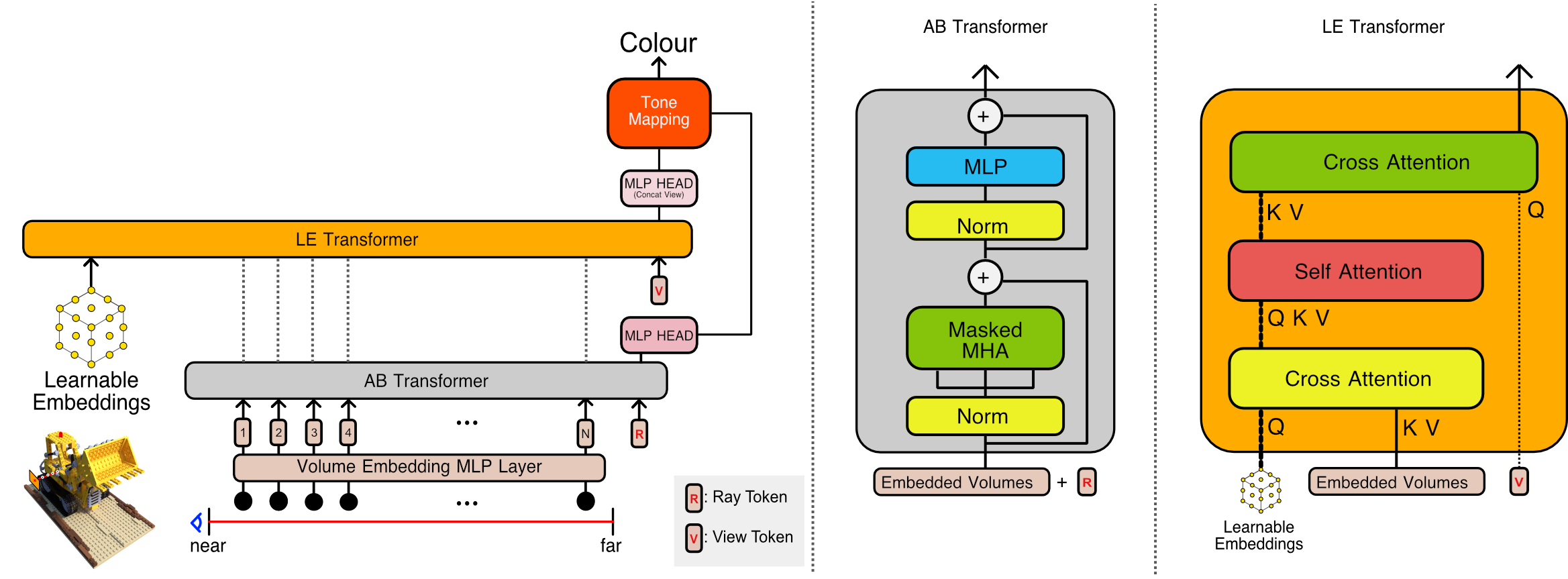}
\caption{A visualisation of ABLE-NeRF. Similar to mip-NeRF, we cast a ray and sample for N conic frustum volumes between the near and far boundary. Each volume passes through a Volume Embedding layer consisting of several layers of MLP. A ray token `R' is appended to the sequence of points before propagating it to the Attention-Based rendering Transformer (AB Transformer) module. After the last transformer layer, the ray token is used to compute a non explicit view-dependent colour. Next, several Learnable Embedding (LE) and a view-dependent token `V' are appended to the sequence of embedded volumes post AB Transformer module before passing to LE Transformer. Within LE Transformer, LE cross-attend to the embedded volumes to memorise static scene information. LE then processes this information with self-attention and a view-dependent token `V' decodes from LE. The final colour is produced by a tone mapping function that takes into account both the colour and view tokens, after the MLP head.}
\label{fig:neuralengine}
\end{figure*}

As aforementioned, applying NeRF's volumetric rendering to accumulate features in the colour space causes the outgoing radiance to be highly sensitive to both opacity \(\sigma\) prediction and the point sampling intervals \(\delta\). Despite the \(\delta\) intervals, the density \(\sigma\) of each point acts as a differential opacity for controlling the accumulated radiance along a ray passing through space \cite{nerf}. As such, NeRF has difficulty predicting the colour of a surface point exhibiting both transmissive and reflective properties at different angles, resulting in a `murky' appearance. ABLE-NeRF addresses this issue by diverging from such physics-based volumetric rendering equation. Instead, we formulate an attention-based network in ABLE-NeRF to determine a ray's colour. These changes allow ABLE-NeRF a flexibility to selectively assign attention weights to points compared to alpha compositing point features (\ref{eqn:weight}) along a ray. We constrained the attention mechanism by introducing masks where frontal points are restricted from attending to rear points. This masking strategy allows us to encode a viewing directional information implicitly. In addition, to capture view-dependent appearance caused by forms of indirect illumination, we incorporate LE as a methodology inspired by light and reflection probes from modern game engines.
\subsection{Attention-based Volumetric Rendering} \label{sec:abvr}

NeRF predicts both  $\sigma$ value and colour of a sampled point. As a consequence, NeRF faces difficulties in predicting a surface that exhibits both translucent and reflective properties at different angles shown in fig.~\ref{fig:drums-1}. Authors of \cite{refnerf} attribute NeRF's inadequacy in predicting an object's specular effects to the difficulty in interpolating between `highly complicated functions' for glossy appearance. We further extend this argument, stating it is even more challenging to interpolate between glossy and translucent appearances of a sampled point that exudes \emph{both} characteristics.

To solve this issue, we can decompose the problem into rendering translucent and reflective surfaces separately. Determining a point's $\sigma$ is equivalent to controlling a point's opacity \cite{nerf}. Therefore, points along a translucent surface should have low $\sigma$ values to describe a low radiance accumulation along a ray. Conversely, for an entirely reflective surface, the points of the reflective surface should have a high $\sigma$ value to indicate a high outgoing radiance. Thus, predicting a point's $\sigma$ is critical in describing its outgoing radiance. However, in NeRF, $\sigma$ is fixed for a point that is either translucent or reflective at different angles. In this scenario, the task of predicting a point's outgoing radiance is left to the viewing directional MLP, which is ill-equipped to do so.

Inspired by the use of volumetric rendering (\ref{eqn:weight}), the weight of a point depends on the weights of itself and the frontal points lying along the same ray. In our work, we apply a transformer model to generate the weights of individual points of the same ray. With this approach, we do not generate $\sigma$ values directly based on the spatial position of a sampled conic frustum of volume \cite{mipnerf}. Instead of assigning weights based on $\sigma$ and $\delta$ as per (\ref{eqn:color}), the importance of a point contributing to a ray's radiance is determined by an attention mechanism.

For a given ray, we sample $N$ number of conic frustums of volumes along it encoded with Integrated Positional Encoding (IPE) described in mip-NeRF. Each conic volume passes through a volume embedding block of four MLP layers to generate a volume embedding $\mathbf{v}^i$, where $i$ denotes the position of conic volume along the ray starting from the camera,  with latent dimensional size of $D$. Similar to ViT \cite{dosovitskiy2020image} and BERT's [class] token \cite{devlin2018bert}, we prepend a ray token $\mathbf{R}$ of the same dimension to the sequence of volume embeddings.  We abuse the notation of sets to describe an input sequence  $\mathbb{Z}_0$, as a set of ray token and the sequence of embedded conic volumes, to the first transformer layer as described in (\ref{eqn:z0}). The subscript notation in $\mathbb{Z}$ is used to denote the number of successive self-attention operations on the set.

In a manner similar to (\ref{eqn:weight}), we utilise a `masking' technique to the limit the attention of volume embeddings solely to those that lie ahead of them along the ray, thereby excluding all others. Specifically, a volume embedding can only attend to the itself, the ray token, and other volume embeddings lying in front of it. This exclusion is represented in the set exclusion shown in (\ref{eqn:xL}), where conic volumes sequenced behind $\mathbf{v}^i$ are excluded from the standard self attention operation. The masking is expressed by setting the scaled-dot product attention to $-\infty$ in the input of softmax, similar to the decoder settings of Transformers, to denote illegal connections \cite {vaswani2017attention}. This `masking' constraint allows us to implicitly encode view-dependent information; zero masking indicating a bi-directional ray, while masking constraints it to being uni-directional. We demonstrate in sec.~\ref{sec:masking} the importance of masking.  No masking is applied to the ray token (\ref{eqn:xR}).

After the final encoder layer L, a single MLP classification head is attached to $\mathbf{R}_L$ to predict the colour of the ray (\ref{eqn:y}). The equations are presented below.


\begin{equation} \label{eqn:z0} 
\mathbb{Z}_0 = \{\mathbf{R},\mathbf{v}^{1}, \mathbf{v}^{2}, ... ,\mathbf{v}^{N} \}
\end{equation} 
\begin{equation} \label{eqn:xL}
\mathbf{v}_{l}^i = \text{Att}(\mathbb{Z}_{l-1}   \backslash  \{\mathbf{v}^{i+1}_{l-1}, ... ,\mathbf{v}^{N}_{l-1}  \}  ) 
\end{equation}
\begin{equation} \label{eqn:xR}
\mathbf{R}_{l} = \text{Att}(\mathbb{Z}_{l-1}  ) 
\end{equation}
\begin{equation} \label{eqn:y}
\mathbf{y} = \text{MLP}(\mathbf{R}_{L})
\end{equation}


\subsection{Hierarchical Volume Sampling with Coarse-Fine Feature Propagation}
We follow the general NeRF rendering strategy in creating two networks: coarse and fine. In NeRF, the coarse network uses \(N_c\) stratified samples as inputs and then re-samples \(N_f=\frac{1}{2}N_c\) points. Next, the fine network uses the total \(N_c+N_f \) points to produce the final colour. Unlike  NeRF, mip-NeRF samples \(N_c=N_f\) conic frustum volumes for each of the coarse and fine networks. The final predicted ray colour uses only \(N_f\) samples for computation, discarding information from the coarse network. In our work, the coarse network also uses \(N_c\) stratified samples. To generate \(N_f=N_c\) samples in our fine network, we sample from the attention weights of the output coarse ray token at state $\mathbf{R}_L^C$ (after $L$ layers attending to all the coarse volume embeddings in the coarse network). Unlike mip-NeRF which discards coarse sample information entirely, we retain this information by reusing coarse ray token as the input fine ray token ($\mathbf{R}_0^F = \mathbf{R}_L^C$) for the fine network. Thus, we retain the ray representation from the coarse network. This approach allows us to avoid the quadratic cost of scaling up to an entire \(N_c+N_f\) samples in every transformer layer of the fine network and only rely on $N_f$ samples.

\subsection{Learnable Embeddings for View-Dependent Appearance}
NeRF's rendering process strictly calculates the radiance of points along a ray. However, the directional MLP is insufficient in computing the specularities of a point. Other NeRF variants attempt to resolve this with some success by predicting a reflection direction of each point \cite{refnerf}. The general rendering equation \cite{kajiya1986rendering} describes how indirect illumination should include multi bounce lighting, where lights reflects off surfaces and shines upon other surfaces. In NeRF's strict rendering ray casting approach, only points on the ray are used for radiance computation. Consequently, NeRF's rendering model can only coarsely approximate direct and indirect illumination using a view direction. We are interested in resolving this issue by capturing the indirect illumination effects radiated by other possible sources. Hence, it is imperative to formulate a query process for external sources \emph{beyond volumes along a ray}. Inspired by game engines' usage of probes, we create LE to store static scene information. These LE serves as a form of memory which allows us to design a secondary branch of attention mechanism as seen in fig.~\ref{fig:neuralengine}.

Like the ViT class token, Learnable Embeddings (LE) in our work are trainable network parameters (memory tokens) used to capture static lighting information by querying from conic frustums in latent space. The iterative training process whereby LE attends to conic volumes in the scene allows the scene lighting information to be encoded as memory. During inference, conic volumes are mapped into latent space via these embeddings and then decoded with a view directional token. In our architecture, the view token is a camera pose Fourier encoded by 16 bands and mapped to the same dimension as LE via a linear layer. 

\subsection{Tone Mapping}
The attention-based rendering backbone outputs the direct illumination exuded by the conic frustum of volumes along the ray. Separately, the cross-attention branch with LE outputs the view dependent illumination of these volumes. In this manner, we prevent the network from over-fitting with this separation. To combine both the outputs, we apply a fixed mapping function to convert linear colours to sRGB, capped to [0,1] as Ref-NeRF \cite{refnerf}.

\section{Experiments}
We implement our model on two datasets; the Blender dataset and Shiny Blender dataset. Similar to mip-Nerf \cite{mipnerf}, we sample conic frustums of volumes along a ray. Our model maintains two networks, coarse and fine. The number of transformer layers, $L$, described in sec.~\ref{sec:abvr} is 2 and 6 for the coarse and fine networks respectively. The coarse network is designed as a lighter network with fewer layers, as its purpose is to generate fine samples, similar to mip-NeRF 360 \cite{mipnerf360} proposal MLP. We sample 192 conic frustums in total, 96 samples in each network, and included 32 LE (shared by coarse and fine networks) to store view-dependent information. The volume embedding module consists of 4 MLP layers, each with 192 hidden units, with ReLU activations. The dimensions of each transformers are set to 192, the same dimension as the volume embedding layers and in the feed-forward layers, the FF hidden unit ratios are set to 1:3. For the Shiny Blender dataset, we set the number of LE to 16, as it contains simpler objects compared to the standard Blender dataset. Optimisation on each scene is trained for 250k iterations.

On each dataset, we evaluate ABLE-NeRF with three commonly used image quality metrics; PSNR, SSIM, LPIPS. A full breakdown of per-scene scores is included in the supplementary materials.

\begin{figure}[t]
\centering
\includegraphics[width=1.0\linewidth]{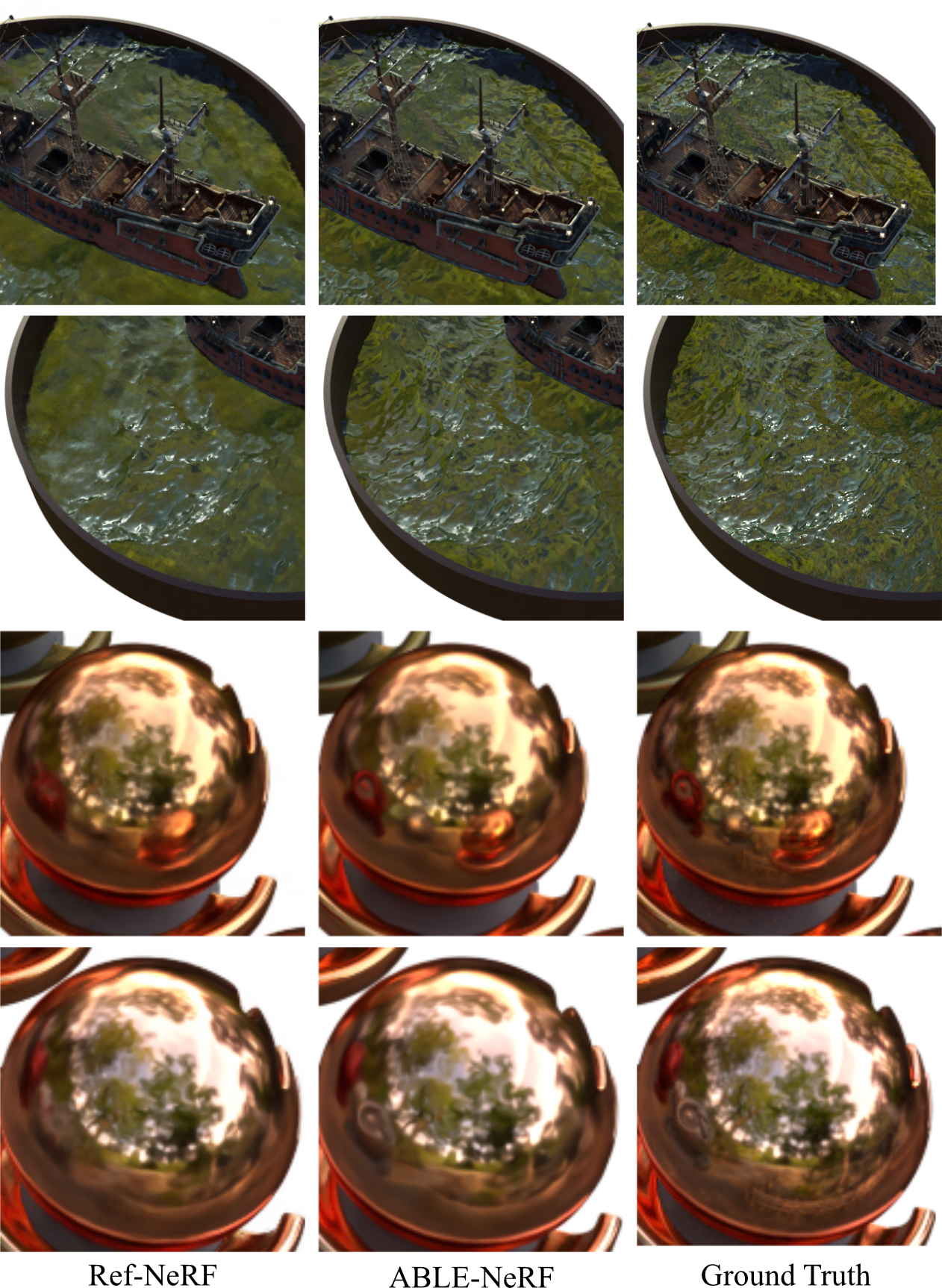}
\caption{ABLE-NeRF significantly improves upon visual realism of highly complex surfaces such as the waves in the Blender ship scene. Furthermore ABLE-NeRF is able to capture intra-scene reflections of neighbouring spheres off glossy sphere in the Blender Materials scene. Top performing NeRF based variant often fail in producing surfaces of complex geometries and challenging view-dependent multi-bounce lighting.}
\label{fig:comparison}
\end{figure}

\subsection{Blender Dataset}
\begin{table}
    \centering
    \small
    \begin{tabular}{ c|c c c } 
     Model & PSNR $\uparrow$ & SSIM $\uparrow$ & LPIPS $\downarrow$ \\ 
     \hline
     PhySG & 20.60 & 0.861 & 0.144\\ 
     VolSDF & 27.96 & 0.932 & 0.096 \\ 
     NSVF & 31.74 & 0.953 & 0.046 \\
     NeRF & 32.38 & 0.957 & 0.043 \\
     Mip-NeRF & 33.09 &  0.961 & 0.043 \\
     Ref-NeRF & \cellcolor{yellow!50}33.99 & \cellcolor{orange!50}0.966 & \cellcolor{orange!50}0.038 \\
     \hline
     Ours, no LE & \cellcolor{orange!50}34.05 & \cellcolor{yellow!50}0.963 & \cellcolor{yellow!50} 0.041 \\
     Ours & \cellcolor{red!40}35.02 & \cellcolor{red!40}0.975 & \cellcolor{red!40}0.035 \\
    
     \hline
    \end{tabular}%
    \caption{Baseline comparisons of ABLE-NeRF and previous approaches on Blender dataset. Results extracted from \cite{refnerf}.}
    \label{tab:blender}

\end{table}

We compare ABLE-NeRF with the latest neural based synthesis network on the standard Blender dataset that originated from NeRF's paper. The results in Table \ref{tab:blender} shows that our work surpasses prior art when compared to the top performing NeRF based method which applies a physics-based volumetric rendering. 

ABLE-NeRF also outperforms prior art qualitatively in rendering photo-realistic appearances of surfaces. As seen in fig.~\ref{fig:comparison}, ABLE-NeRF renders compelling visuals of highly complex surfaces in the Blender Ship scene where the surfaces of the waves resemble the ground truth more closely compared to Ref-NeRF. In the Materials scene, ABLE-NeRF produces reflections of intra-scene objects, attributed to the use of LE, which captures multi-bounce lighting effects. The appearance of reflections of spheres off another neighbouring sphere (reflections of reflections) is clearer compared to standard ray-casting approaches of NeRF. This highlights the importance of maintaining LE to capture indirect lighting effects.

\subsection{Shiny Blender Dataset}

\begin{table}
    \centering
    \small
    \begin{tabular}{ c|c c c } 
     Model & PSNR $\uparrow$ & SSIM $\uparrow$ & LPIPS $\downarrow$ \\ 
     \hline
     PhySG & 26.21 & 0.921 & 0.121 \\ 
     Mip-NeRF & 29.21 & \cellcolor{yellow!50}0.942 & \cellcolor{yellow!50}0.092 \\
     Ref-NeRF (no pred. normals) & \cellcolor{yellow!50} 30.91 & 0.936 & 0.105 \\
     Ref-NeRF & \cellcolor{red!40}35.96 & \cellcolor{orange!50}0.967 & \cellcolor{red!40}0.058 \\
     \hline
     Ours & \cellcolor{orange!50}33.88 & \cellcolor{red!40}0.969 & \cellcolor{orange!50}0.075 \\
    
     \hline
    \end{tabular}
    \caption{Baseline comparisons of ABLE-NeRF and previous approaches on Shiny Blender dataset. Results extracted from \cite{refnerf}.}
    \label{tab:shinyblender}

\end{table}

\begin{figure}[t]
\centering
\includegraphics[width=1.0\linewidth]{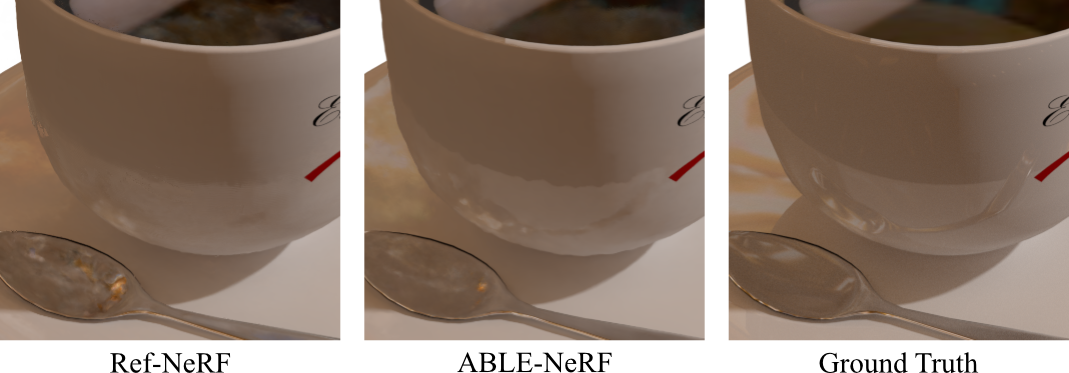}
\caption{ABLE-NeRF is able render intra-scene surface reflections better than Ref-NeRF. As shown in the Shiny Blender Coffee scene, the reflection of the teaspoon on the side of the cup appears more apparent than Ref-NeRF. }
\label{fig:cofee}
\end{figure}

\begin{figure}[t]
\centering
\includegraphics[width=1.0\linewidth]{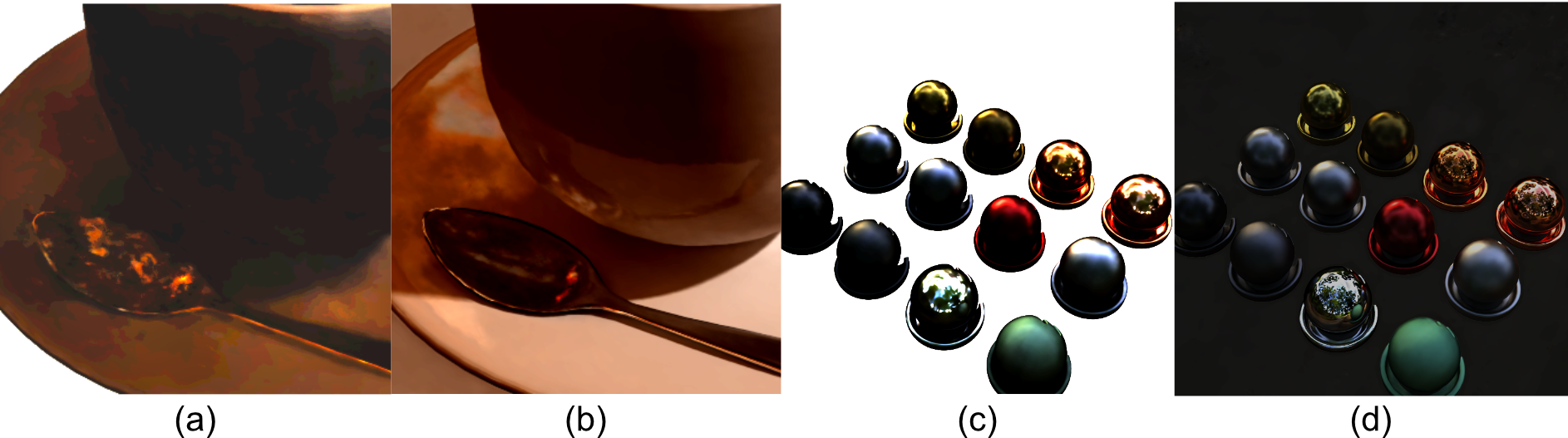}
\caption{Prior to tone-mapping, (a) and (c) are outputs from the AB Transformer while (b) and (d) are outputs from the LE Transformer.}
\label{fig:sep}
\end{figure}
Compared to the Blender dataset by NeRF \cite{nerf}, the Shiny Blender dataset by Ref-NeRF \cite{refnerf} contains objects with more glossy surfaces. It is important to note that the Shiny Blender dataset mostly consists of singular objects with simple geometries, where the surfaces are smooth and rounded. As a result, Ref-NeRF outperforms ABLE-NeRF in terms of PSNR and LPIPS since the normals for such objects are easier to predict, compared to the complex surfaces in the standard Blender dataset of NeRF. For example, for a smooth rounded `Shiny Ball', Ref-NeRF outperforms ABLE-NeRF due to its simpler geometry. However, for a more complex surface such as the `Toaster', ABLE-NeRF outperforms Ref-NeRF. We display Ref-NeRF ablation study results with no normal predictions to support our case. Without normal predictions, ABLE-NeRF surpasses Ref-NeRF by a wide margin.

It is worth highlighting to readers that ABLE-NeRF excels at capturing intra-scene reflections of surfaces caused by multi-bounce lighting, which are highly complex scenes. In such scenarios, the reflections of objects interact with other objects within the scene. In fig.~\ref{fig:cofee}, we show reflections of the teaspoon off the cup in the `Coffee' scene is rendered closely to ground truth. Ref-NeRF fails to capture such intra-scene reflections compared to ABLE-NeRF. The intra-scene reflection due to multi-bounce lighting is well captured as shown in fig.~\ref{fig:sep}.

\section{Architectural Analysis}

\subsection{Masking Strategy} \label{sec:masking}

\begin{figure}[t]
\centering
\includegraphics[width=1.0\linewidth]{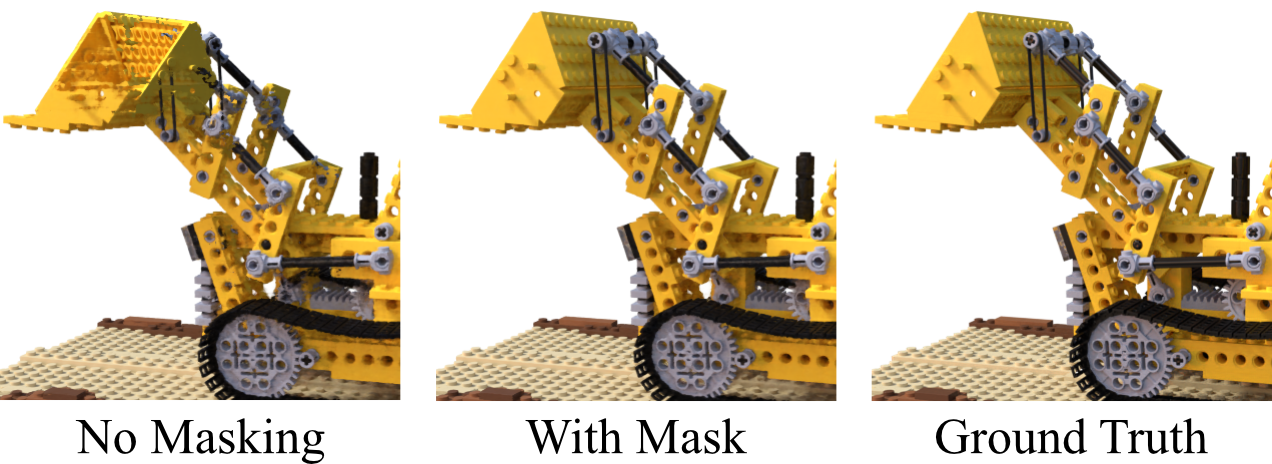}
\caption{Here we show the importance of masking rear points from frontal point along a camera ray. Without mask, a bi-directional ray is implied causing the network to have difficulty rendering the object's surface accurately. With the masking strategy, we enable transformers to mimic a volumetric rendering strategy and also implicitly encode a view directional information.}
\label{fig:masking}
\end{figure}

The masking strategy is imperative in allowing transformers to model volumetric rendering as an end to end deep learning based approach. Without masks, the model would render inaccurate surfaces as seen in fig.~\ref{fig:masking}. By including masks, we implicitly encode a uni-directional ray versus a non-masking bidirectional ray as shown in the figure. We have attempted to introduce a uni-directional ray information without masks by appending a volume's position on a ray. However, this attempt is ineffective compared to our original masking strategy.

\subsection{Learnable Embeddings}

\begin{figure*}[ht]
\centering
\includegraphics[width=1.0\linewidth]{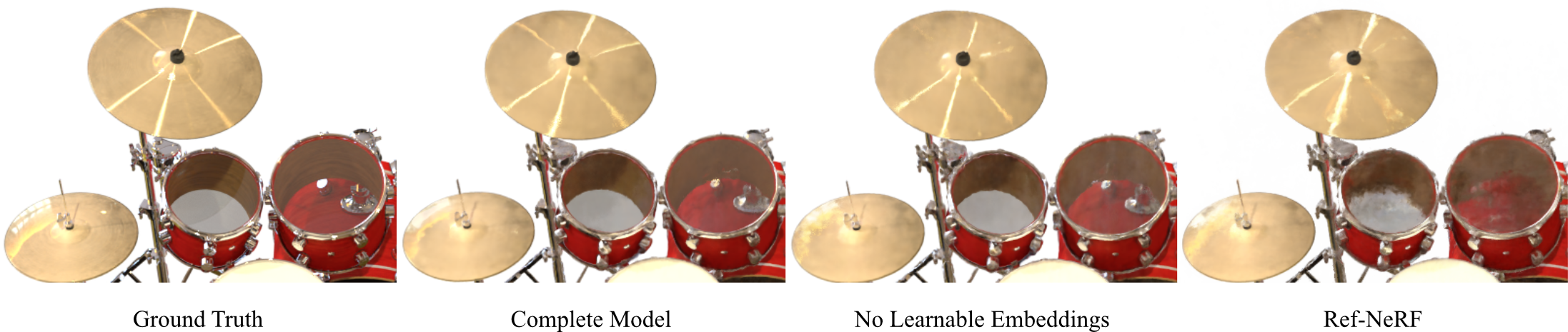}
\caption{We visualise how ABLE-NeRF benefits from the use of transformers for volumetric rendering and also the inclusion of LE. Observe the specularities of the right and centre cymbals. Removing LE causes the model to fail in capturing specular effects effectively on the cymbals. Even without LE, the basic backbone of using transformers as a deep learning VR-based approach allows us to render the translucent portions of the drums more accurately compared to NeRF based approaches.}
\label{fig:neuralprobes}
\end{figure*}

\textbf{Learnable Embeddings Inclusion} We validate our model design choices by performing an ablation study on the Blender dataset. In this setup, we exclude LE and compute the final ray colour using only the ray token. The results are included in Table \ref{tab:blender}. Without LE, the model performs comparatively with Ref-NeRF. By including LE, ABLE-NeRF has the flexibility to attend to embeddings not constrained to the line of ray. Evidently, we demonstrate in fig.~\ref{fig:neuralprobes} that LE allows our model to capture better view-dependent specularities.

\begin{figure}[t]
\centering
\includegraphics[width=1.0\linewidth]{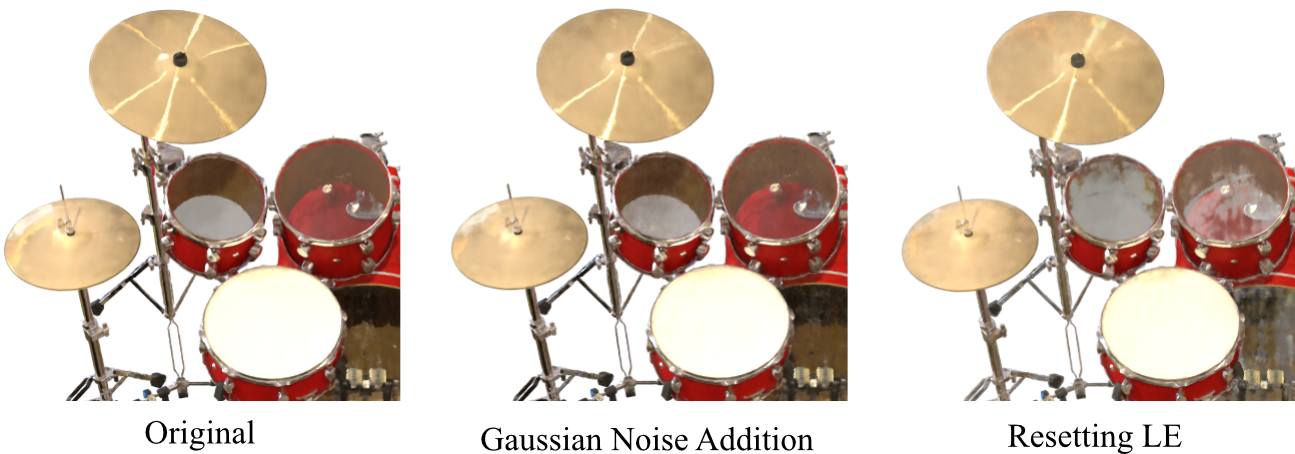}
\caption{As we corrupt the weights of LE with additive Gaussian noise, we observe that the view-dependent surfaces of the drum scene changes. As we continue to destroy the weights of LE by setting it to zero, we corrupt the specularities and transparencies of the drums. Observe that the specularity in left cymbal of the uttermost right figure is completely eradicated. Diffuse surfaces largely remain unchanged perceptually.}
\label{fig:perturb}
\end{figure}

\textbf{Perturbating Learnable Embedding} We formally described LE as a form of memory for a static scene. As a form of analysis to understand what LE are effectively memorising, we perturbed these memory by adding Gaussian noise to corrupt these memory tokens. Lastly, we wiped the memory of LE, collapsing all memory into a single entity, by setting its weights to zero. With reference to fig.~\ref{fig:perturb}, observe that the diffuse surfaces remains perceptually unchanged while view-dependent surfaces with specularities and translucency are affected. This analysis offers us insights to how LE could be modified to edit scenes dependent on viewing direction for future work.

\begin{figure}[t]
\centering
\includegraphics[width=1.0\linewidth]{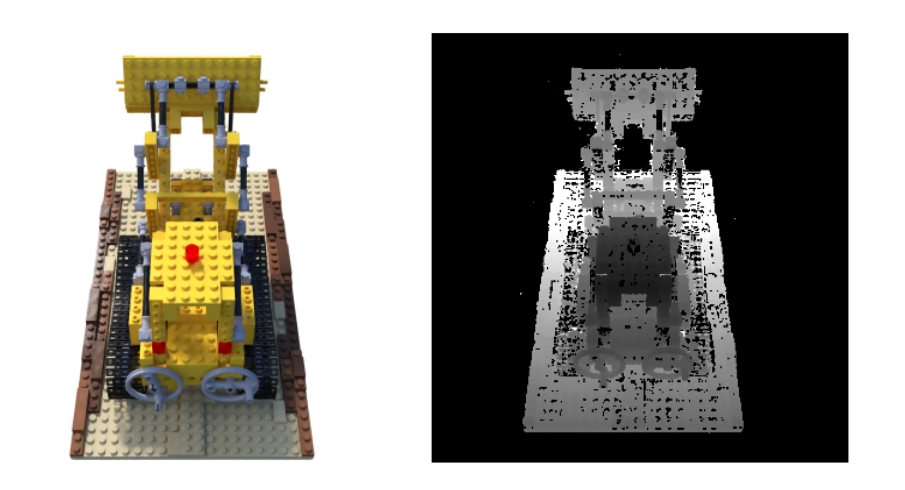}
\caption{We extract the per-volume attention from the ray token. By selecting the highest attention weight to the volume, we are able to plot a depth map based on the distance traversed to that conic volume frustum along the ray from the camera origin.}
\label{fig:depth}
\end{figure}

\subsection{Attention Maps as Depth Maps}

As the ray token selectively assigns attention weights to conic volume frustums along the ray, the volume with the highest attention weight could imply a surface of the object in the scene. With the attention map, we plot a depth map based on the distance traversed from the camera origin to the volume with the highest attention weight. Fig.~\ref{fig:depth} illustrates the capability of ABLE-NeRF in generating depth maps from attention weights.


\section{Conclusion}

We have highlighted the general issues of NeRF-based rendering methods that use physics-based volumetric rendering, which cause difficulties in rendering complex surfaces that exhibits both translucent and specular appearances. Our model, ABLE-NeRF, ameliorates such issues by applying a deep learning-based method using masking on transformers to learn a physics-based volumetric rendering method. With the attention weights generated by transformers, we can re-sample a 3D space effectively with visual content and output a depth map from an attention map. Lastly, we have included Learnable Embeddings as a form of memorisation framework to capture view-dependent lighting effects in latent space and allow the view angle token to query these LE beyond a ray for accurate view-dependent visuals. These contributions allow ABLE-NeRF to significantly improve upon prior art in novel view synthesis. We believe that our work paves the way forward in rendering accurate visuals of complex objects and scenes, as well as hinting at the potential for new scene editing methods by reprogramming LE.
Our code is available at \url{https://github.com/TangZJ/able-nerf}.
\\
\textbf{Acknowledgements}
This study is supported under the RIE2020 Industry Alignment Fund – Industry Collaboration Projects (IAF-ICP) Funding Initiative, as well as cash and in-kind contribution from the industry partner(s).

{\small
\bibliographystyle{ieee_fullname}
\bibliography{egbib}
}


\clearpage
\appendix

\captionsetup{labelformat=AppendixTables}
\setcounter{table}{0}

\section{Optimisation Details}
Our implementation is based on PyTorch Lightning. We optimise our model with Adam and we set the learning rate as $5 \times 10^{-4}$ which is annealed log-linearly to $1 \times 10^{-4}$, with a warm-up phase of 1250 iterations. We set the hyperparameters as $\beta_1 = 0.9$ and $\beta_2 = 0.999$.  
Each scene in the Blender and Shiny Blender dataset takes 1.5 days to train on 8 Tesla v100 GPUs with a total batch size of 8192 rays. To render an image on a single Tesla v100 GPU takes 110s. 

\section{Blender Dataset Details}
We report full breakdown of individual scenes in the Blender Dataset in Tables A.\ref{tab:psnr-blender}, A.\ref{tab:ssim-blender}, A.\ref{tab:lpips-blender}.

We compare our results with the rendered test set images from Ref-NeRF\footnote{We thank Dor Verbin for providing us with the test images from Ref-NeRF model output}.

\begin{table}[ht]
    \centering
    \resizebox{\columnwidth}{!}{%

    \begin{tabular}{ l | *{8}{c}}
      & chair & lego & materials & mic & hotdog & ficus & drums & ship\\ 
     \hline
     PhySG \cite{zhang2021physg} & 24.00 & 20.19 & 18.86 & 22.33 & 24.08 & 19.02 & 20.99 & 15.35 \\
     VolSDF \cite{yariv2021volume} & 30.57 & 29.46 & 29.13 & 30.53 & 35.11 & 22.91 & 20.43 & 25.51 \\
     Mip-NeRF \cite{mipnerf} & 35.12 & 35.92 & 30.62 & 36.76 & 37.34 & 33.19 & 25.36 & 30.52 \\
     Ref-NeRF \cite{refnerf} &  \cellcolor{orange!50}35.83 &  \cellcolor{yellow!50}36.25 &  \cellcolor{orange!50}35.41 &  \cellcolor{orange!50}36.76 &  \cellcolor{yellow!50}37.72 &  \cellcolor{yellow!50}33.91 &  \cellcolor{yellow!50}25.79 &  \cellcolor{yellow!50}30.28 \\
     \hline
     Ours,No LE &  \cellcolor{yellow!50}35.76 &  \cellcolor{orange!50}36.62 & \cellcolor{yellow!50}34.57 &  \cellcolor{yellow!50}35.90 &  \cellcolor{orange!50}38.68 &  \cellcolor{orange!50}34.28 &  \cellcolor{orange!50}25.98 &  \cellcolor{orange!50}30.60 \\
     Ours &  \cellcolor{red!40}36.25 &  \cellcolor{red!40}38.03 &  \cellcolor{red!40}35.46 &  \cellcolor{red!40}37.11 &  \cellcolor{red!40}39.07 &  \cellcolor{red!40}35.69 &  \cellcolor{red!40} 26.84 &  \cellcolor{red!40}31.75  \\
     \hline
    \end{tabular}%
    }
    \caption{Per-scene test set PSNRs on Blender Dataset. Results retrieved from \cite{refnerf}.}
    \label{tab:psnr-blender}

\end{table}

\begin{table}[ht]
    \centering
    \resizebox{\columnwidth}{!}{%

    \begin{tabular}{ l | *{8}{c}}
      & chair & lego & materials & mic & hotdog & ficus & drums & ship \\ 
     \hline
     PhySG \cite{zhang2021physg} &  0.898 & 0.821 & 0.838 & 0.933 & 0.912 & 0.873 & 0.884 & 0.727 \\ 
     VolSDF \cite{yariv2021volume} & 0.949 & 0.951 & 0.954 & 0.969 & 0.972 & 0.929 & 0.893 & 0.842 \\
     Mip-NeRF \cite{mipnerf} & 0.981 & 0.980 & 0.959 & 0.992 & 0.982 & 0.980 & 0.933 & 0.885 \\
     Ref-NeRF \cite{refnerf} & \cellcolor{orange!50}0.984 & \cellcolor{yellow!50}0.981 & \cellcolor{orange!50}0.983 & \cellcolor{orange!50}0.992 & \cellcolor{yellow!50}0.984 & \cellcolor{yellow!50}0.983 & \cellcolor{yellow!50}0.937 & \cellcolor{yellow!50}0.880 \\
     \hline
     Ours, No LE & \cellcolor{yellow!50}0.941 & \cellcolor{orange!50}0.985 & 0\cellcolor{orange!50}.983 & \cellcolor{yellow!50}0.991 & \cellcolor{orange!50}0.987 & \cellcolor{orange!50}0.986 & \cellcolor{orange!50}0.945 & \cellcolor{orange!50}0.893 \\
     Ours & \cellcolor{red!40}0.987 & \cellcolor{red!40}0.987 & \cellcolor{red!40}0.985 & \cellcolor{red!40}0.993 & \cellcolor{red!40}0.988 & \cellcolor{red!40}0.989 & \cellcolor{red!40}0.951 & \cellcolor{red!40}0.919 \\

    \end{tabular}%
    }
    \caption{Per-scene test set SSIMs on Blender Dataset. Results retrieved from \cite{refnerf}.}
    \label{tab:ssim-blender}

\end{table}

\begin{table}[ht]
    \centering
    \resizebox{\columnwidth}{!}{%

    \begin{tabular}{ l| *{8}{c}}

 f 
      & chair & lego & materials & mic & hotdog & ficus & drums & ship \\ 
     \hline
     PhySG \cite{zhang2021physg} & 0.093 & 0.172 & 0.142 & 0.082 & 0.117 & 0.112 & 0.113 & 0.322 \\
     VolSDF \cite{yariv2021volume} & 0.056 & 0.054 & 0.048 & 0.191 & 0.043 & 0.068 & 0.119 & 0.191  \\
     Mip-NeRF \cite{mipnerf} & \cellcolor{yellow!50}0.020 & 0.018 & 0.040 & \cellcolor{orange!50}0.008 & 0.026 & 0.021 & \cellcolor{yellow!50}0.064 & \cellcolor{orange!50}0.135 \\
     Ref-NeRF \cite{refnerf} &  \cellcolor{red!40}0.017 & \cellcolor{orange!50}0.018 & \cellcolor{red!40}0.022 & \cellcolor{red!40}0.007 & \cellcolor{orange!50}0.022 & \cellcolor{yellow!50}0.019 & \cellcolor{orange!50}0.059 & \cellcolor{yellow!50}0.139 \\
     \hline
     Ours, No LE & 0.021 & \cellcolor{orange!50}0.018 & \cellcolor{yellow!50}0.031 & 0.010 & \cellcolor{yellow!50}0.024 & \cellcolor{orange!50}0.017 & 0.065 & 0.147 \\
     Ours & \cellcolor{orange!50}0.018 & \cellcolor{red!40}0.015 & \cellcolor{orange!50}0.029 & \cellcolor{orange!50}0.008 & \cellcolor{red!40}0.021 & \cellcolor{red!40}0.014 & \cellcolor{red!40}0.057 & \cellcolor{red!40}0.122 \\

    \end{tabular}%
    }
    \caption{Per-scene test set LPIPS on Blender Dataset. Results retrieved from \cite{refnerf}.}
    \label{tab:lpips-blender}

\end{table}

\section{Shiny Blender Dataset Details}

We report full breakdown of individual scenes in the Shiny Blender Dataset in Tables A.\ref{tab:psnr-sblender}, A.\ref{tab:ssim-sblender}, A.\ref{tab:lpips-sblender}.

We compare our results with the rendered test set images from Ref-NeRF.
\begin{table}[ht]
    \centering
    \resizebox{\columnwidth}{!}{%

    \begin{tabular}{ l| *{6}{c}} & teapot & toaster & car & ball & coffee & helmet  \\
    \hline
    PhySG \cite{zhang2021physg} & 35.83 & 18.59 & 24.40 & 27.24 & 23.71 & 27.51  \\
    Mip-NeRF \cite{mipnerf} & 46.00 & 22.37 & 26.50 & 25.94 & 30.36 & 27.39 \\
    Ref-NeRF, no pred. normals \cite{refnerf} & \cellcolor{yellow!50}47.09 &\cellcolor{yellow!50} 23.32 & \cellcolor{yellow!50}27.19 & \cellcolor{yellow!50}26.09 & \cellcolor{yellow!50}31.79 &\cellcolor{orange!50} 30.54 \\
    Ref-NeRF \cite{refnerf} & \cellcolor{red!40}47.90 & \cellcolor{orange!50}25.70 & \cellcolor{red!40}30.82 & \cellcolor{red!40}47.46 & \cellcolor{red!40}34.21 & \cellcolor{yellow!50}29.68 \\
    \hline
    Ours & \cellcolor{orange!50}47.30 & \cellcolor{red!40}26.52 & \cellcolor{orange!50}28.76 & \cellcolor{orange!50}36.62 & \cellcolor{orange!50}33.01 & \cellcolor{red!40}31.04 \\ 

    \end{tabular}%
    }
    \caption{Per-scene test set PSNRs on Shiny Blender Dataset. Results retrieved from \cite{refnerf}.}
    \label{tab:psnr-sblender}

\end{table}

\begin{table}[ht]
    \centering
    \resizebox{\columnwidth}{!}{%

    \begin{tabular}{ l| *{6}{c}} & teapot & toaster & car & ball & coffee & helmet  \\
    \hline
    PhySG \cite{zhang2021physg} & 0.990 & 0.805 & 0.910 &\cellcolor{yellow!50}  0.947 & 0.922 & 0.953  \\ 
    Mip-NeRF \cite{mipnerf} & 0.997 & 0.891 & 0.922 & 0.935 & 0.966 & 0.939 \\
    Ref-NeRF, no pred. normals \cite{refnerf} &  \cellcolor{orange!50}0.997 & \cellcolor{yellow!50}0.898 & \cellcolor{yellow!50}0.926 & 0.865 & \cellcolor{yellow!50}0.967 & \cellcolor{orange!50}0.962 \\
    Ref-NeRF \cite{refnerf} & \cellcolor{red!40}0.998 & \cellcolor{orange!50}0.922 & \cellcolor{red!40}0.955 & \cellcolor{red!40}0.995 & \cellcolor{orange!50}0.974 & \cellcolor{yellow!50}0.958  \\
    \hline
    Ours & \cellcolor{red!40}0.998 & \cellcolor{red!40}0.949 & \cellcolor{orange!50}0.942 &\cellcolor{orange!50} 0.984 & \cellcolor{red!40}0.975 & \cellcolor{red!40}0.968 \\ 

    \end{tabular}%
    }
    \caption{Per-scene test set SSIMs on Shiny Blender Dataset. Results retrieved from \cite{refnerf}.}
    \label{tab:ssim-sblender}

\end{table}

\begin{table}[ht]
    \centering
    \resizebox{\columnwidth}{!}{%

    \begin{tabular}{ l| *{6}{c}} & teapot & toaster & car & ball & coffee & helmet  \\
    \hline
    PhySG \cite{zhang2021physg} & 0.022 & 0.194 & 0.091 & 0.179 & 0.150 & 0.089  \\ 
    Mip-NeRF \cite{mipnerf} &  0.008 & \cellcolor{yellow!50}0.123 & \cellcolor{yellow!50}0.059 & \cellcolor{yellow!50}0.168 & \cellcolor{orange!50}0.086 & 0.108  \\
    Ref-NeRF, no pred. normals \cite{refnerf} &  \cellcolor{orange!50}0.006 &0.134 & 0.064 & 0.272 & \cellcolor{yellow!50}0.087 & \cellcolor{red!40}0.068 \\
    Ref-NeRF \cite{refnerf} & \cellcolor{red!40} 0.004 & \cellcolor{orange!50}0.095 & \cellcolor{red!40}0.041 &\cellcolor{red!40} 0.059 & \cellcolor{red!40}0.078 &\cellcolor{orange!50} 0.075 \\
    \hline
    Ours &\cellcolor{yellow!50} 0.007 & \cellcolor{red!40}0.092 & \cellcolor{orange!50}0.052 & \cellcolor{orange!50}0.107 & 0.114 & \cellcolor{yellow!50}0.080 \\ 

    \end{tabular}%
    }
    \caption{Per-scene test set LPIPS on Shiny Blender Dataset. Results retrieved from \cite{refnerf}.}
    \label{tab:lpips-sblender}

\end{table}

\end{document}